\RequirePackage{fix-cm}
\documentclass[smallextended]{svjour3}       
\smartqed  
\usepackage{graphicx}
\usepackage[nocompress]{cite}
\usepackage{graphicx}
\usepackage{verbatim}
\usepackage{caption}

\usepackage{cite}

\usepackage{graphicx}

\usepackage{diagbox} %
\usepackage{algorithm}
\usepackage{algorithmicx}
\usepackage{algpseudocode}

\algnewcommand\algorithmicforeach{\textbf{for each}}
\algdef{S}[FOR]{ForEach}[1]{\algorithmicforeach\ #1\ \algorithmicdo}

\usepackage{subfigure}
\usepackage{amsmath}
\usepackage{enumerate}
\begin{document}

\title{Active Collaboration in Relative Observation for Multi-agent Visual SLAM based on Deep Q Network}


\author{Zhaoyi Pei \and
  Piaosong Hao \and
  Meixiang Quan \and
  Muhammad Zuhair Qadir \and
  Guo Li
}


\institute{Zhaoyi Pei \at Department of Computer Science and Technology,
  Harbin Institute of Technology, Harbin, China\\
  \email{peizhaoyi@stu.hit.edu.cn}           
  \and
  Songhao Piao \at Department of Computer Science and Technology,
  Harbin Institute of Technology, Harbin, China \\ 
  \email{piaosh@hit.edu.cn}
  \and
  Meixiang Quan \at
  Department of Computer Science and Technology,
  Harbin Institute of Technology, Harbin, China \\ 
  \email{15b903042@hit.edu.cn}
  \and
  Muhammad Zuhair Qadir \at
  Department of Computer Science and Technology,
  Harbin Institute of Technology, Harbin, China \\ 
  \email{mzuhairqadir@gmail.com}
  \and
  Guo Li \at
  Department of Computer Science and Technology,
  Harbin Institute of Technology, Harbin, China \\
  \email{14B903020@hit.edu.cn}
}

\date{Received: date / Accepted: date}

\maketitle
\begin{abstract}
  This paper proposes a unique active relative localization mechanism for multi-agent Simultaneous Localization and Mapping(SLAM),in which a agent to be observed are considered as a task, which is performed by others assisting that agent by relative observation. A task allocation algorithm based on deep reinforcement learning are proposed for this mechanism. Each agent can choose whether to localize other agents or to continue independent SLAM on it own initiative. By this way, the process of each agent SLAM will be interacted by the collaboration. Firstly, based on the characteristics of ORBSLAM, a unique observation function which models the whole MAS is obtained. Secondly, a novel type of Deep Q network(DQN) called MAS-DQN is deployed to learn correspondence between Q Value and state-action pair, abstract representation of agents in MAS are learned in the process of collaboration among agents. Finally, each agent must act with a certain degree of freedom according to MAS-DQN. The simulation results of comparative experiments prove that this mechanism improves the efficiency of cooperation in the process of multi-agent SLAM.
\end{abstract}
\section{Introduction}
\paragraph A critical task of a mobile robot is to determine its pose (position and orientation) under the given environment map, which is also the basis of other tasks. However, the environment map does not exist at the beginning. When the mobile robots enter an unknown environment, they need to construct the map (3d point cloud map, topological map, 2D map) to satisfy their tasks through its own sensors, and at the same time determine their own localization in the map, this is called simultaneous localization and mapping(SLAM) problem. Mapping and localization are mutually complementary. In order to build accurate maps, it is necessary for the robot to have an accurate estimation of its pose, Vice versa,localization also requires established high-quality maps, it is the difficulty of SLAM. Visual SLAM refers to the SLAM that only the camera acts as the only visual sensor. Compared with the traditional laser, the visual sensors can obtain more abundant information and can be deployed on mobile robots at very low cost. 
\paragraph Multi-agent system(MAS) refers to that agents in one system collaborate and negotiate with others to accomplish goals that can not be accomplished by a single agent or to increase the efficiency of task execution. Each agent's ability, attributes and structure are different, which provides a great space for multi-agent collaboration. MAS have attracted many researchers to investigate the applicability of it in many pertinent areas closely related to human livelihood and industry such as security system\cite{Rahman}, exploration and rescue\cite{Azimi}, surveillance\cite{Du}, humanitarian assistance\cite{Saaidia}, environment protection \cite{Espina}\cite{Shkurti} health care\cite{Rodriguez} \cite{Jemaa}. Traditional SLAM has been considered as an independent behavior. With the continuous application of MAS, multi-agent SLAM has become the focus of scholars' research.
\paragraph Multi-agent SLAM can be divided into two types: off-line mode and on-line mode. Off-line mode means that each agent's slam process is independent and the results are fused together after all agents stop. On-line mode refers to that agents can cooperate with each other in slam process, such as cooperative loop closure detection and cooperative localization, cooperative map fusion, etc. Relative orientations in MAS are important on account of that it is an basis for combining maps built with data from two agents. The difficulty of online multi-agent SLAM is how to cooperate in the best way under the condition that each agent has different attributes with different states. This issue is particularly prominent in the relative localization between robots. For instance, the localization ability of wheeled robot is stronger than that of UAV because of the odometer, but the pose  of UAV is more abundant, which leads to its strong mapping ability. If there are two different kinds of robots in the system at the same time, An appropriate collaboration model will make full use of the advantages of both.\cite{Surmann} This relative localization is controlled manually. If we want to optimize the efficiency of automatic cooperative positioning between agents, each agent needs to perceive all the states and their attributes related to SLAM to determine whether it has the optimal condition to observe the agent in need of assistance. But in most cases, the task-related attributes and states of robots are not quantifiable which brings difficulties to collaboration.
\paragraph In order to solve the above problem of SLAM of agents in MAS, this paper proposes a cooperative mechanism applied to multi-agent SLAM, which can learn the main attributes related to each agent and extracts the state feature vectors of all agents in MAS based on ORBSLAM\cite{ORBSLAM}, each agent will determine its specific behavior outputted by MAS-DQN. In the experiment section, we use model of isomorphic robots with different attributes to simulate various robots, and the validity of our algorithm is proved.
\paragraph The paper is organized as follows: in the section 2, related work about coalition formation for MAS and multi-agent SLAM is discussed. In the section 3, the DQN and "Rainbow" which we applied to drive MAS-DQN are introduced. The structure of MAS-DQN including the novel observation function and reward function is described in the section 4. Section 5 introduces relative observation between agents based on the proposed mechanism.The effectiveness of the algorithm is discussed in Chapters 6. The complete algorithm of the approach is shown in section 7.
\section{Related Work}
\paragraph task allocation has always been a hot topic in MAS research. Task allocation in MAS(MRTA) can be formulated as an optimal assignment problem where the goal is to optimally assign a set of agents to a set of tasks in such a way that optimizes the overall system performance under a condition of satisfying a set of constraints.According to team organizational paradigm, MRTA can be divided into two categories: centralized approaches and distributed approaches.
\paragraph There is a central agent that monitors and assigns tasks to other agents in a centralized MAS. agents who are assigned tasks need to send all the information they obtain to that central agent. They need to keep communication with each other at all times. The center-based algorithm is very general for solving the MRTA problem as a result that monitors can directly control each agent.\cite{Al-Yafi} ,Coltin, B\cite{Coltin} proposed a centralized approach for mobile robot task allocation in hybrid wireless sensor networks. In literature\cite{Liu}, A centralized multi-robot task allocation for industrial plant inspection by using A* and genetic algorithm is introduced. Higuera, J\cite{Higuera} proposed fair division-based MRTA approach as another centralized algorithm to handle the problem of allocating a single global task among heterogeneous robots.
\paragraph In configuration of distributed approaches, Agents do not need to report to a central agent, and they do not need to be online at all times. They can freely communicate with other agents and assign tasks through negotiation with different demands. This type of algorithm is also widely used to solve MRTA problem. Farinelli A, Iocchi L proposed a d istributed on-line dynamic task assignment for multi-robot patrolling.\cite{Farinelli} literature \cite{Luo} proposed distributed algorithm for constrained multi-robot task assignment for grouped tasks. Various ambient assisted living (AAL) technologies have been proposed for MRTA in Healthcare Facilities.\cite{Das} Emotional robots model use artificial emotions and AI to endow the robot with emotional factors, which makes the robot more intelligent and adjust its behavior choice through the emotional mechanism. The introduction of emotional and personal factors improves the diversity and autonomy of robots.\cite{Jafari}
\paragraph literature\cite{Howe} presented the theoretical basis of relative localization between agents by using prior knowledge in maps. literature\cite{Zhou} proposed a mechanism for relative localization in which agents try to observe each other in-time if they have a chance with unknown initial correspondence. In the both cases methods should provide rules to correct local world state for single observer using data from another one. literature \cite{Zkucur} establishes the relationship between relative observation and map fusion.
\paragraph Many different types of representation of maps such as gragh-based map\cite{Lazaro}\cite{Grisetti},Landmarks with covariance matrix\cite{Lauritzen} grid-based map \cite{Eliazar}\cite{Huleski} are also used to improve the efficiency of both.
\paragraph Most of the relative observation approaches serve for map fusion. In the mechanism proposed by this paper, the data from relative observations between agents is also applied for optimizing the pose graph of the observed agents which improves the efficiency of data utilization.
\section{DQN and Rainbow}
Q-learning is a off-policy learning method in reinforcement learning. It uses Q-Table to store the cumulative value of each state-action pair. According to Bellman equation, when the strategy of maximizing Q value is adopted in every step, the $Q(s_t,a_t)$ can be calculated as equation \ref{bellman}.
\begin{equation}
  Q(s_t,a_t)=r_t+\gamma*(argmax_{a'}Q(s_{t+1},a'))
  \label{bellman}
\end{equation}
where the $Q(s_t,a_t)$ refers to the cumulative discounted reward when agent's action is $a_t$ at states $s_t$. $s_t$ is the state on time $t$ and $s_{t+1}$ is the state on time $t+1$,$a_t$ is the action adapted by agent at time $t$. The state of the system is transformed to $s_t+1$ from $s_t$ for the action $a_t$ taken by agent.$r_t$ is the immediate reward for the transition from $s_t$ to $s_{t+1}$.$\gamma$ is discounted factor.
When the state and action space are high-dimensional or continuous, the maintenance of Q-Table $Q(s,a)$ will become unrealistic. Therefore, with the problem of updating Q-Table transformed into a problem of fitting function, the basic idea of DQN is to use the neural network to make the Q function $Q_\theta(s,a)$ approach the optimal Q value by updating the parameter $\theta$ which refers to all weights of the deep neural network. The specific sequence of DQN is shown in algorithm 2.
\begin{algorithm}
  \begin{algorithmic}[1]
    \label{DQN}
    \caption{DQN}
    \caption{The training process of DQN}
    \State Initialize replay memory D to capacity N
    \State Initialize $Q$-function with random weights $\theta$
    \State Initialize $Q$-function with weights $\overline{\theta}=\theta$
    \For {episode = 1:M}
    \State Initialize sequence $s_1=\{x_1\}$ and preprocessed sequence $\phi_1=\phi(s_1)$
    \For {t = 1:T}
    \State select $a_t$ then observe reward $r_t$ and $x_{t+1}$
    \State Processed $\phi_{t+1}=\phi(x_{t+1})$ and store transition ($\phi_t,a_t,r_t,\phi_{t+1}$) in D
    \State Sample mini-batch of transition ($\phi_j,a_j,r_j,\phi_{j+1}$)
    \State Set
    \begin{equation}  
      y_j=\left\{  
      \begin{array}{lr}
        r_j                                                 & if episode \, terminates \, at \, step \, j+1 \\
        r_j+\gamma max_{a'}Q(\phi_{j+1},a',\overline\theta) & otherwise 
      \end{array}
      \right.  
    \end{equation}
    \State Perform a gradient descent step on ($y_j-Q(\phi_j,a_j,\theta)$) w.r.t. network parameter $\theta$
    \State Every $C$ steps reset $\overline(\theta)=\theta$
    \EndFor
    \State $i \to i+1$
    \EndFor
  \end{algorithmic}
\end{algorithm}
where the $x$ is the state of the system that has not been purified by observation function, However, the combination of deep learning(DL) and reinforcement learning(RL) will inevitably lead to some problems, such as:
\begin{enumerate}
  \item DL needs a large number of labeled samples for supervised learning; RL only has the reward of state-action pairs as return value.
  \item The sample of DL is independent, and the state of RL is correlative.
  \item The distribution of DL targets is fixed; the distribution of RL is always changing, that is, it is difficult to reproduce the situation that was trained before.
  \item Previous studies have shown that the use of non-linear networks to represent value functions is unstable.
\end{enumerate}
This paper used "Rainbow" to build the training model of MAS-DQN. Rainbow\cite{Rainbow} is an integrated DQN. It mainly integrates the following algorithms to try to solve the above problems:
\begin{enumerate}
  \item Double Q-Learning constructs two neural network with the same structure but different parameters: Behavior Network and Target Network. When updating the model, the best action at t+1 time is selected by Behavior Network, then the target value of the optimal action at t+1 time is estimated by Target Network. By decoupling these two steps, we can reduce the impact of Q-Learning method on overestimation of value.
  \item  Priority Replay Buffer make the model choose more samples that can improve the model to use the replay memory more efficiently.
  \item Dueling DQN decomposes the value function into two parts, making the model easier to train and expressing more valuable information.
  \item Distributional DQN changes the output of the model from computational value expectation to computational value distribution, so that the model can learn more valuable information.
  \item Noisy DQN adds some exploratory ability to the model by adding noise to the parameters, which is more controllable.
  \item In order to overcome the weakness of slow learning speed in Q-learning earlier stage, Multi-step Learning uses more steps of reward, so that the target value can be estimated more accurately in the early stage of training, thus speeding up the training speed.
\end{enumerate}
Rainbow model integrates the advantages of the above models which proves the possibility of integrating the above features successfully.
\section{The Network for Propose Mechanism}
\subsection{Observation Function for MAS-DQN based on ORBSLAM}
In deep reinforcement learning, the local information obtained by an agent is called observation. Unlike previous deep reinforcement learning algorithms, the observation function of the proposed mechanism models all agents and determines the behavior of each agent at the same time instead of for only one agent. In this section, ORBSLAM feature vector and output of the observation function is introduced.
\subsubsection{ORBSLAM2 Feature Vector}
\paragraph ORBSLAM is one of the most commonly used system among modern visual SLAM frameworks. It runs in three threads: In tracking thread, the system tracks motion of the platform by using tracked feature points; Local mapping thread is used for constructing global map and optimizing local map, Loop closure thread take responsibility for detecting loop when agent revisited to an already mapped area, which can prevent increase of the accumulative errors.
\paragraph ORBSLAM is deployed on each agent independently to simultaneous construct the map. In our system, the state of the environment consist of all agents pose estimation state which is mainly determined by three items: map points,key frames and loop closure detection. Some variables related to the three items are used to compose the state vector of the agent called ORBSLAM feature vector whose members are introduced next.
\begin{enumerate}[1]
  \item The first of component of the ORBSLAM feature vector is variable related to the map point. The map point is constructed from the matched point features. We include the number of map points observed from current frame in feature vector.
  \item The second of component of the feature vector are variables related to the key frames which are the selected frames according to strict requirements to ensure that they contain sufficient map information. Some key frames proved to contain insufficiently accurate map information will be gradually removed. In order to estimate the content of map information at nearby locations which play a decisive role in pose estimation, the number of the generated new key frames and that of the eliminated old key frames from our last sampling time in the state vector are included in the state vector.
  \item The third of component of the feature vector is variable related to the result of loop closure. Loop closure detection is the problem that a mobile agent determines whether itself has returned to a previously visited location. If a closed-loop is detected, then it will impose a strong constraint on global optimization. In feature vector. we add the time interval of the frame from last loop closure detected to measure the impact of loop closure to the current frame, 
  \item  Another critical factor is the collection of distances from the agent to other agents. The longer the distance from the agent to target agent is, the bigger difficult to it to assist the target. The distance is calculated by D* algorithm. The characteristic of D* algorithm is that when the terrain between the target point and the source point is known, the shortest path and the shortest distance can be obtained. If not fully known, the shortest path in the known terrain and the estimated distance between the target point and the source point can be output.An orbslam feature vector contains m-1 (m is the number of agents) such distances, representing the distance between the agent and the other m-1 agents respectively.
\end{enumerate}
An ordered arrangement of all the above variables is an ORBSLAM feature vector. On the next subsubsection the observation function base on the orbslam feature vector will be introduced.
\subsubsection{The Observation Function}
The output of the observation function represents the perception of the overall slam state of the whole multi-agent system. The orbslam feature vectors of all agents at the same time form an observation frame. Considering the continuity of the slam process, the previous states should be taken into count when the system makes macro decision for each agent. So the output of the observation function should include the current frame and the fixed number of preceding frames. These frames consist of an observation frame set, which eventually flattens into a vector as the output of the observation function. The specific steps of the observation function are shown in algorithm 3.
\begin{algorithm}
  \begin{algorithmic}[1]
    \label{Observation}
    \caption{Observation function of the proposed algorithm}
    \Require $O$ : Set of ORBSLAM feature vector.
    \Require $t$ : Current timestamp.
    \Ensure $X$: Global perception of the current state of the environment.
    \Function {ObservationFunction}{$O$,$t$}
    \For {$i = 1 : n$}
    \For {$j =1 : m$}
    \If {t-i+1 >0}
    \State x.extend($O(t-i+1,j)$) 
    \Else
    \State x.extend($O(1,j)$)
    \EndIf
    \State $j \to j+1$
    \EndFor
    \State $i \to i+1$
    \EndFor
    \EndFunction
  \end{algorithmic}
\end{algorithm}
While the parameter n is the number of observation frames, m is the total number of agents in the system, and $O(i,j)$ is the orbslam2 feature vector of Agent j at time i.
\subsection{Reward Function}
The reward function is generally used to reflect the immediate benefit of an agent reaching a new state through action. In our proposed algorithm, the transition error is used to measure the reward, which is shown in the equation \ref{loss}.
\begin{equation}
  loss = \sqrt{(x'-x)^2+(y'-y)^2+(z'-z)^2}
  \label{loss}
\end{equation}

In order to embody the collaboration between agents, the reward function also adds the influence of agent's target on his decision-making. The final reward function of agent is as equation\ref{reward} :
\begin{equation}
  r(i) = -(\mu loss(i)+\epsilon(1-\mu)(loss_k(t)-loss_{k-1}(t)))
  \label{reward}
\end{equation}
where $loss_{k}(i)$ represents the $loss$ of agent i at time k. $(loss_k(t) -loss_{k-1}(t))$ denotes the contribution of agent $i$ to his target $t$ to assist. as the selfishness of the agent,$\mu $ is used to control the frequency of agent reorganization, $\epsilon$ depends on the result of the assistance, and its value, such as the equation \ref{epsilon}, will be discussed in detail in the next subsection.
\begin{equation}  
  \left\{  
  \begin{array}{lr}
    \epsilon=1  & When\, the\, target\,agent\, is\, successfully\, assisted\,by\,this\,agent\                       \\  
    \epsilon=0  & When\,the\, target\, is\, itself\, or\, the\, position\, for\, collaboration\, is\, not\, reached \\  
    \epsilon=-1 & When\, the\, collaboration\, is\, failed                  
  \end{array}
  \right.  
  \label{epsilon}
\end{equation}
\subsubsection{The output formation of the proposed DQN}
There are m*(m+1) neurons in the output layer, and the output of $(i-1)*(m+1)+j$th neurons represents the cumulative reward value of the assistant behavior to the jth agent of the ith agent learned by DQN. The $j$th is 0 means that no action will be taken for the agent which is waiting for the assistance of other agents. SLAM is carried out independently for agent i when $j=i$. Each agent will be given an order with the maximum cumulative reward $argmax_{a}Q(s,a)$ with a certain probability. How agents can assist other agents will be described in the next section. The structure of the whole neural network is shown in the following figure \ref{MAS-DQN}.
\begin{figure}
  \includegraphics[width=0.8\linewidth]{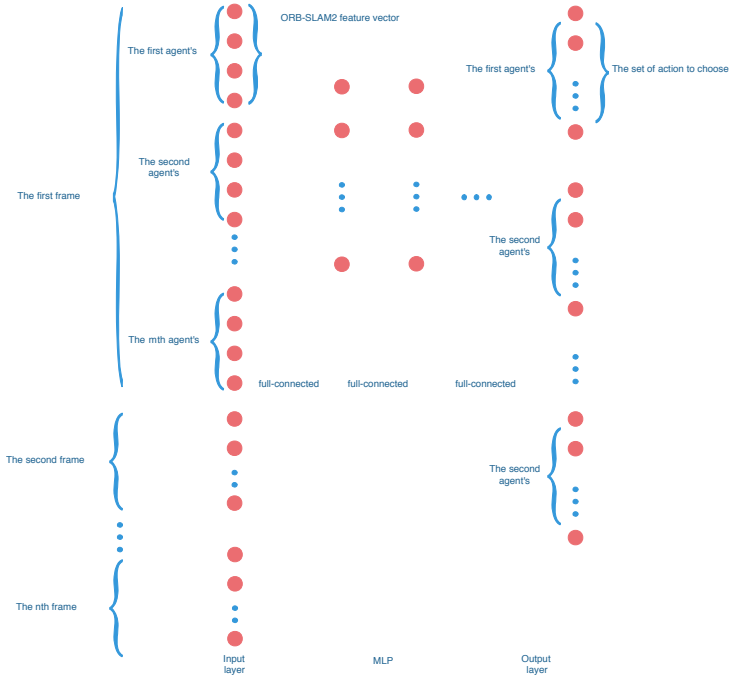}
  \centering
  \caption{structure of MAS-DQN}
  \label{MAS-DQN}
\end{figure}
\section{Relative Observation between Agents in the proposed Mechanism}

\begin{figure}
  \includegraphics[width=0.8\linewidth]{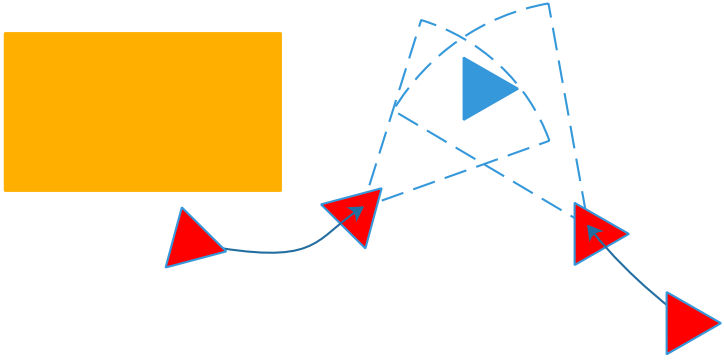}
  \centering
  \caption{Assistance in MAS based on relative observation}
  \label{assistgraph}
\end{figure}

In the proposed mechanism, agents execute the order given by MAS-DQN with each others through mutual observation. As shown in figure \ref{assistgraph}, red agents are assigned to assist blue agents. First, they shall reach the predetermined observation position. If all the predetermined positions can not be reached, the observation will be judged as a failure, agent obtains the pose of the target through the relative localization of the target agent. The obtained position and posture will provide a strong constraint for the target agent to optimize the pose graph. In another word, it is used to improve the accuracy of correcting pose. This observation actually has the same effect as the loop closure detection. The way of calculating the pose of the target agent is introduced consequently.
\paragraph Assume that agent i observes a feature point $p ^{i j}$ of target agent j whose coordinate in agent i's camera coordinate system is $p ^{i j}_c $, then the coordinate $p ^{i j}_w$ in the world coordinate frame based on agent i's pose can be obtained by equation 2.
\begin{equation}
  p^{ij}_w=R_i*p^{ij}_c+t_i
\end{equation}
Where the $R_i$ refers to agent i's rotation matrix which converts camera coordinates into world coordinates and $t^i$ refers to translation matrix $R$ and $t$ together constitute the pose of robot camera, because the world coordinates of one feature point are the consistent. the relationship between the world coordinates of point P and the coordinates of point P in the camera coordinate system of Agent J is shown in equation 3.
\begin{equation}
  p_w=R_j*p'_c+t_j
\end{equation}
Where $R_j$ and $T_j$ is the pose of agent j, $p'_c$ is the coordinate of feature point $p$ in agent j's 3D model. Because there may be multiple agents observing agent $j$ at the same time, $p_w$ may have multiple values, which makes it impossible to obtain the pose of Agent $j$ directly. Non-linear optimization method can be used to continuously approximate the true value of pose. Firstly, the error of pose estimation is defined as:
\begin{equation}
  e=\frac{1}{2}\Sigma_{i}\Sigma_{k}||p^{ij}_k-(R'_j*p'_k+t'_j)||_2^2
\end{equation}
Where $p ^{i j}_k$ is the coordinate of Kth point of Agent j observed by agent i. $R'$, $t'$ is the estimated value of agent $j's$ pose matrix at this moment.$R'$,$t'$ can be expressed by Lie algebra $exp (\ xi)$, then the optimization of the target function can be expressed as:
\begin{equation}
  min_{\xi}(\frac{1}{2}\Sigma_{i}\Sigma_{k}||p^{ij}_k-exp(\xi)||_2^2)
\end{equation}
The derivative of error term with respect to pose can be obtained by using Lie algebraic perturbation model, and the value of $R'$,$t'$ can be obtained iteratively when $e$ is the minimum value by using non-linear optimization methods such as the gradient descent method.
\section{Effectiveness Proof of the Proposed Algorithm}
\begin{figure}
  \includegraphics[width=0.8\linewidth]{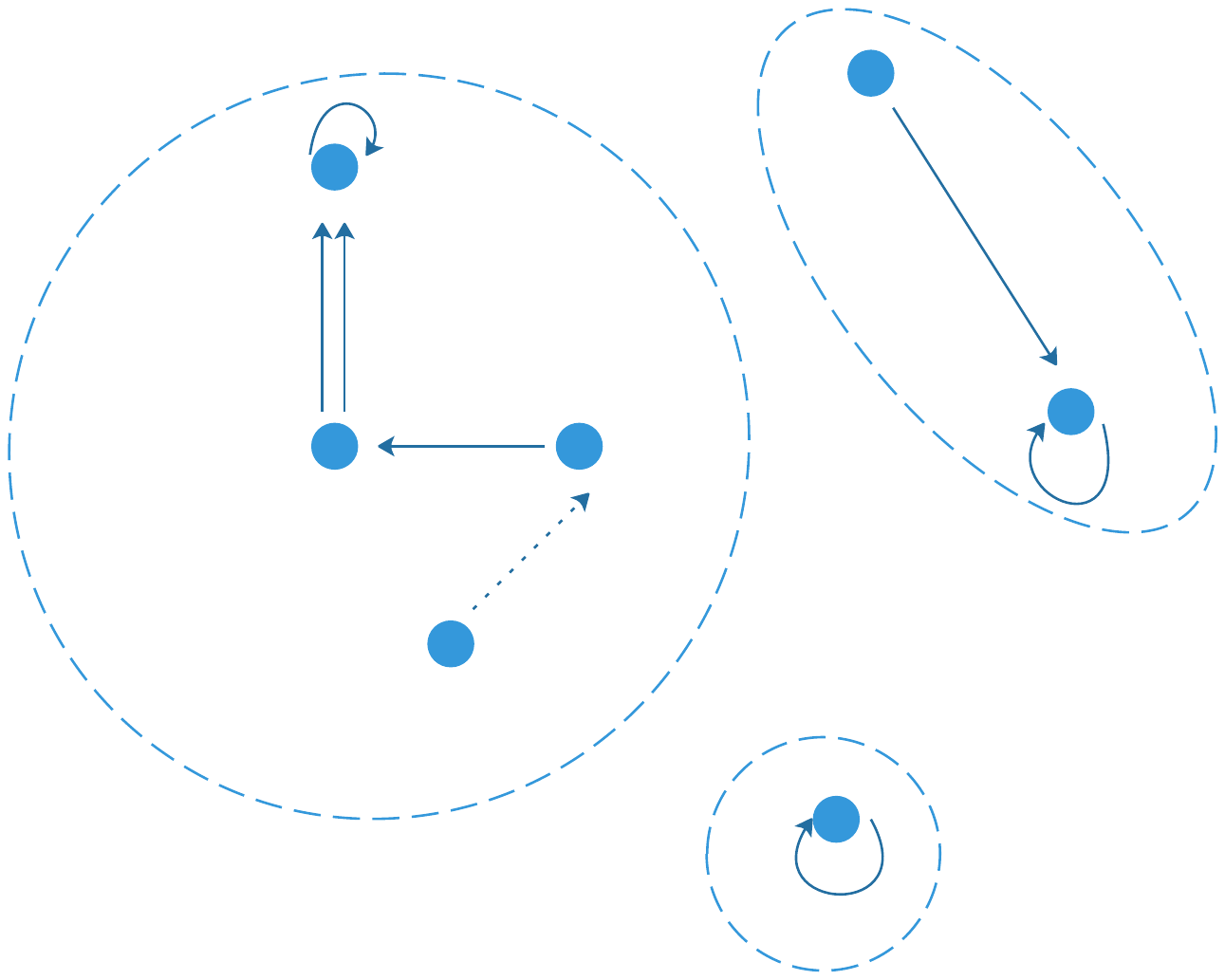}
  \centering
  \caption{Gragh of coordination}
  \label{coordination}
\end{figure}

In this part, we will prove the effectiveness of our algorithm. That is to give the demonstration that after experience replay, the proposed DQN can make macro-decision to improve the efficiency of the whole multi-agent system for each agent after training. We use the utility theory to model the performance of the whole MAS. As shown in the previous section, the effect of SLAM of a single agent $x$ can be regards the immediate utility value of $x$ which is associated with $loss(x)$ :
\begin{equation}
  U(x) = -loss(x) 
\end{equation}
The expectation of the immediate utility value of a non-cooperative multi-agent system is as follows:
\begin{equation}
  E(U(X))=\frac{\Sigma_{x \in X}u(x)}{|X|}
\end{equation}
The calculation of that in collaborative MAS is relatively complex since the effect of collaboration between agents on the overall system utility value should be considered, The graph theory is introduced to calculate the overall utility value of our proposed algorithm. An agent will be regard as a node in the graph and a directed edge from the itself to the target will be established as shown in figure \ref{coordination} . If the collaboration is not executed at that time, the edge is the solid line. If the collaboration is just completed,the edge is a double-line edge. If the collaboration fails, the line is a dashed line, and if there is no collaboration object, the arrow points to itself.The whole multi-agent system can be divided into several maximally connected subgraphs, as shown in the figure \ref{coordination}, where one subgraph is a group which can represent the collaboration relationship between agents. The total utility value of a group $U(g)$ be represented as equation \ref{utility}. 
\begin{equation}
  u(g)_{x \in g} = (u(x)+c(x,x'))
  \label{utility}
\end{equation}
where $x'$ is the target of $x$ and $C(x, x')$ is the effect of assistance $(loss_k(t) -loss_{k-1}(t))$ .
If x belongs to $g$, then x must not belong to any subgraph of $G-g$.The utility of MAS is represented as equation \ref{uti}:
\begin{equation}
  E(u(X))=\frac{\Sigma_{g \in G}U(g)}{|x|} =\frac{\Sigma_{x \in X}r(x)}{|X|}
  \label{uti}
\end{equation}
In summery, to calculate the cumulative reward of all agents is to calculate the total cumulative utility value. MAS-DQN will gradually learn the cumulative utility function $U(S,A(X))$. Where $S$ and $A(X)$ respectively refer to the state space and action space of MAS.

\paragraph The last layer of MAS-DQN is the decision-making layer. All the layers before the decision layer are called feature extraction layer. The feature extraction layer is actually a high-level abstraction of ORBSLAM feature vector. The weights connected with the decision-making level determine the discrepancy of different decision-making of agents. That is, in the face of the same extracted high-level characteristics, different action produce different cumulative reward. This situation is caused by the different attributes between agents, which DQN learns through the cooperation between agents.
\section{Sequences of the proposed mechanism}
The implementation steps of the the proposed overall cooperation mechanism for multi-agent SLAM are shown in algorithm 4.
\begin{algorithm}
  \small
  \scriptsize
  \caption{Complete algorithm}
  \begin{algorithmic}[1]
    \State initialize the MAS-DQN and agents
    \For {$t=1:T$}
    \State Broadcast(requestInfo)
    \State waitingforResponse()
    \ForEach {$x \in X$}
    \State x.response($x.p_w,x.localmap,x.\phi_{t},x.assistantStatus$)
    \EndFor
    \State getAllResponse()
    \State $globelmap \to mergeMap(localMaps[])$
    \State broadcast(globalMap)
    \State generateObservation()
    \State caculateReward()
    \State feed MAS-DQN with $\phi$ and get every agent's action $a$
    \State broadcast(a) for each $x \in X$
    \If {$t > 1$}
    \State $T \to getTransition(\phi_{t-1},a_{t-1},r_t,\phi_{t})$
    \State addExperienceReplyBuffer(T)
    \If {|ReplyBuffer|>Minimum quantity}
    \State MAS-DQN.train(ReplyBuffer)
    \EndIf
    \EndIf
    \ForEach {$x \ in X$}
    \State $x.target \to x.getAction(a)$
    \If {$x.target = x $}
    \State x.randomMove()
    \ElsIf {$x.target = None$}
    \State x.stop()
    \Else
    \State $x.path \to calculatepath(x.target)$
    \If {$x.target_t \neq x.target_{t-1} $}
    \State $x.life \to 0$
    \Else
    \State $x.life \to x.life + 1$
    \If {x.life = Life}
    \State $x.assistantStatus \to fail $
    \EndIf
    \EndIf
    \State x.oneStep(x.path)
    \If {x.onPose()}
    \State x.observe(x.target)
    \If {x.get(x.target)}
    \State x.sendPose(x.target ,pose)
    \State $x.assistantStatus \to successful$
    \State broadcast(assistanceCompleted(x.target))
    \Else 
    \State $x.assistantStatus \to fail$
    \EndIf
    \Else 
    \If {another agent completed assistance to x.target}
    \State $x.assistantStatus \to fail$
    \Else
    \State $x.assistantStatus \to approaching $
    \EndIf
    \EndIf
    \EndIf
    \EndFor
    \EndFor
  \end{algorithmic}
\end{algorithm}
\paragraph The system consists of mobile agent and organizer which MAS-DQN is deployed on. The following is a detailed explanation of this algorithm: Initialize the whole algorithm and all variables.(01) Organizer broadcasts the requirements for synchronizing agent information and waits for the reply of all the agents.(03-04) Each agent respond organizer with their world coordinates, built local maps, ORBSLAM feature vectors, and assistance status.(05-07) the Local maps are not the final form of maps. They are only 2D maps that provide information for agent global navigation. After the organizer obtains  the information returned by all agents(08), the first thing to do for it is to merge the local maps of all agents and feedback the global maps to agents.(09-10) The second mission is to get the output of the observation function according to the set of ORBSLAM feature vector of all agents.(11) Next, The immediate reward of each agent is calculated.(12)Finally, the output is imported to MAS-DQN to get the action of each agent which will be broadcasted.(13-14) $T(\phi_{t-1},a_{t-1},r_t,\phi_{t})$ is added to experience reply buffer.(15-16) After obtaining enough sample information, MAS-DQN will be trained.(17-21) Agent x deconde its own goals from the organizer's instructions $a$.(23) If the target is itself $x$, $x$ will SLAM independently. (24-25) If x is not assigned a target,it will stop to wait for the assistance.(26-27) If x get one target,it  will follow the calculated path to approach his intended location and update his assistance status.(28-54) $Life$ is a constant,which is set up to prevent assistance from going on indefinitely.(30-36) This loop will continue until time runs out.(02-57)
\section{Experiment}
\begin{figure}
  \includegraphics[width=0.8\linewidth]{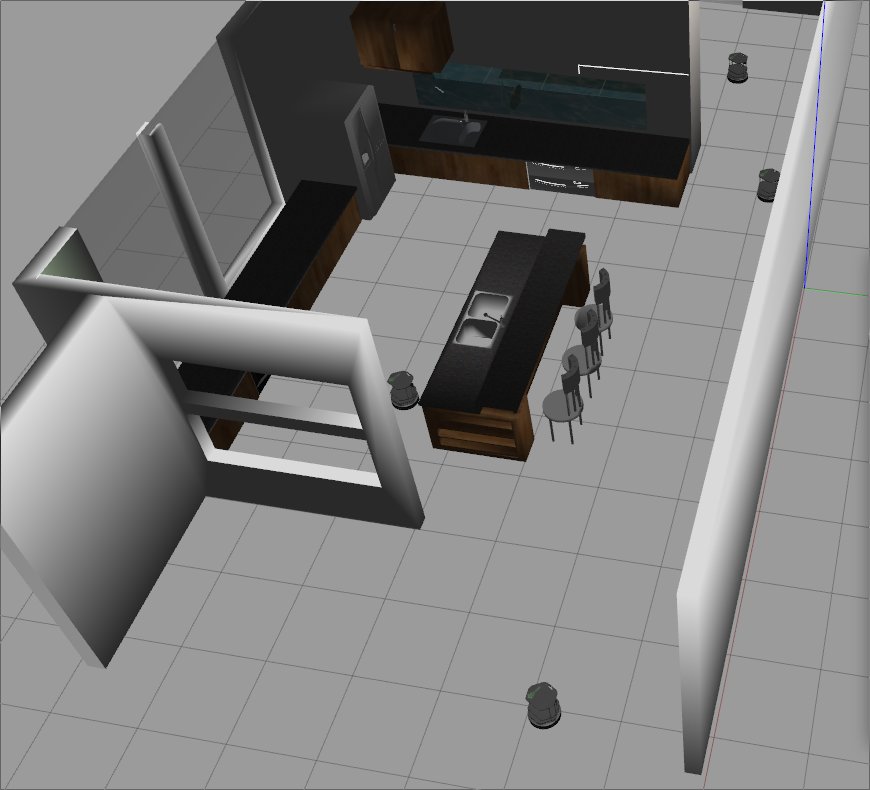}
  \centering
  \caption{experimental world }
  \label{experimentworld}
\end{figure}

The simulation experiments introduced in this section is to evaluate the performance of the proposed mechanism and MAS-DQN with different parameters. The experiment was implemented on the Robot Operating System(ROS) platform which can be used to simulate the real physical environment. We build one simulation mobile agent with the ’telobot’ model which equips “Kinect” as vision sensor. The initial pose of each robot is known and orbslam is deployed independently on each robot. The experimental world is shown as figure \ref{experimentworld}.
\paragraph The main indicators to measure the effect of SLAM in MAS are transition RMSE, orientation RMSE of the tracks of agents. Although agents in experimental environments have the same model, we can embody the differences of their attributes by changing their ability values related to motion and vision sampling. Their maximum angular acceleration, linear acceleration, maximum angular velocity and maximum linear velocity obey normal distribution $N(\mu,\sigma_1)$, in which $\sigma_1$ is a variable, the performance of the algorithms can be tested when the value of agents' attributes vary with different degrees. When it comes to visual sensors, we add noise that obeys the normal distribution $N(\mu,\sigma_2)$ to the images captured by visual sensor, and the variance of the camera noise of different agents $\sigma_2$ still obeys the normal distribution $N(1,\sigma_1)$.
\paragraph We set up four experiments to verify the effectiveness of our algorithm. The following are the details of all experiments and the corresponding experimental results.The experiments have been done with $\sigma_1=0.2$. The time step is set as 0.5 seconds. All the comparative experiments were done after the convergence of training process for MAS-DQN.    

\begin{figure}[H]\centering
  \includegraphics[width=0.45\linewidth]{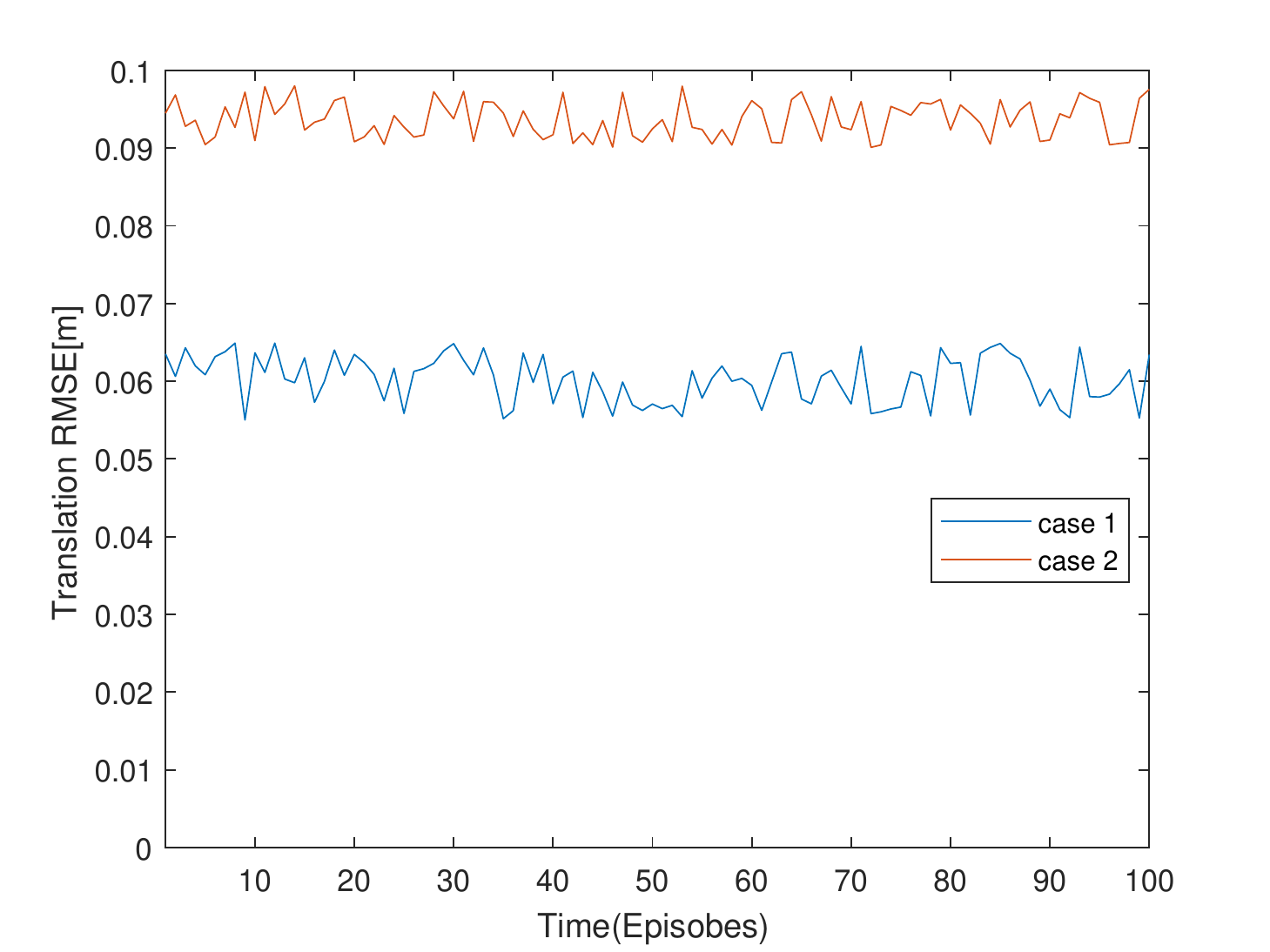}
  \includegraphics[width=0.45\linewidth]{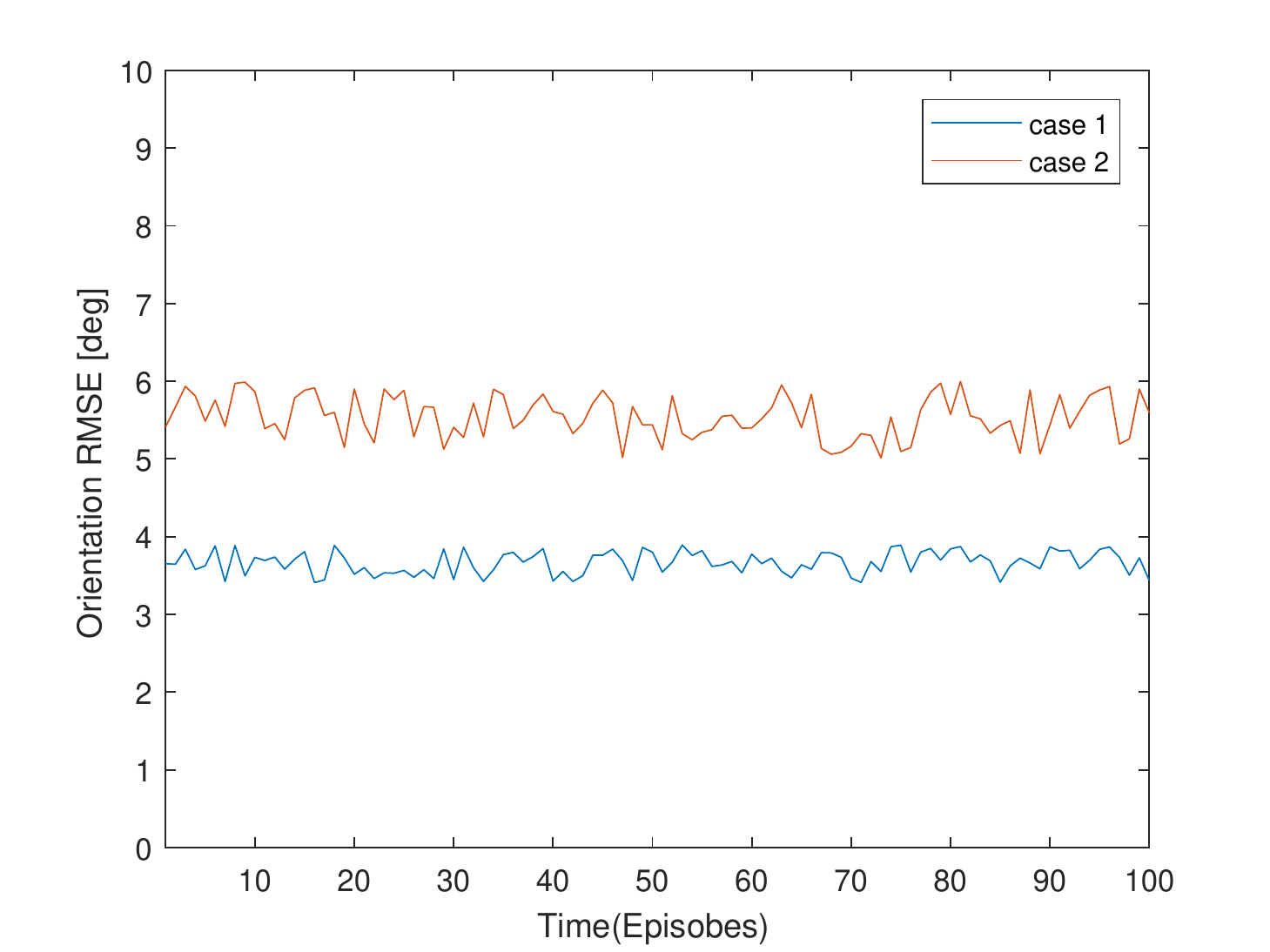}
  \caption{Result of experiment 1}
  \label{withoutcoorperation}
\end{figure}

\paragraph Experiment 1: This experiment shows the role of collaboration in multi-agent SLAM. The results are shown in figure \ref{withoutcoorperation} in which the two algorithms for comparison are explained as follows:
\begin{itemize}
  \item case1: The proposed collaboration mechanism based MAS-DQN.
  \item case2: Each agent SLAM independently without any collaboration.
\end{itemize}

The proposed algorithm reduce transition RMSE(m) by 35.68\% orientation
position RMSE(deg) by 33.86\% compared with the mechanism without coordination

\paragraph Experiment 2:The main purpose of this experiment is to compare mechanism that agents collaborate with each other without considering attributes and states. The experimental results are shown in the figure \ref{withoutcoorperation}.
which illustrate that the proposed algorithm reduce transition RMSE by 28.33\% and orientation RMSE by 33.09\%.
\begin{itemize}
  \item Case 1: The proposed collaboration mechanism based MAS-DQN.
  \item Case 2: Relative localization based on cluster matching algorithm without distance IR in the environment\cite{Rashid}.
\end{itemize}
\begin{figure}[H]\centering
  \includegraphics[width=0.45\linewidth]{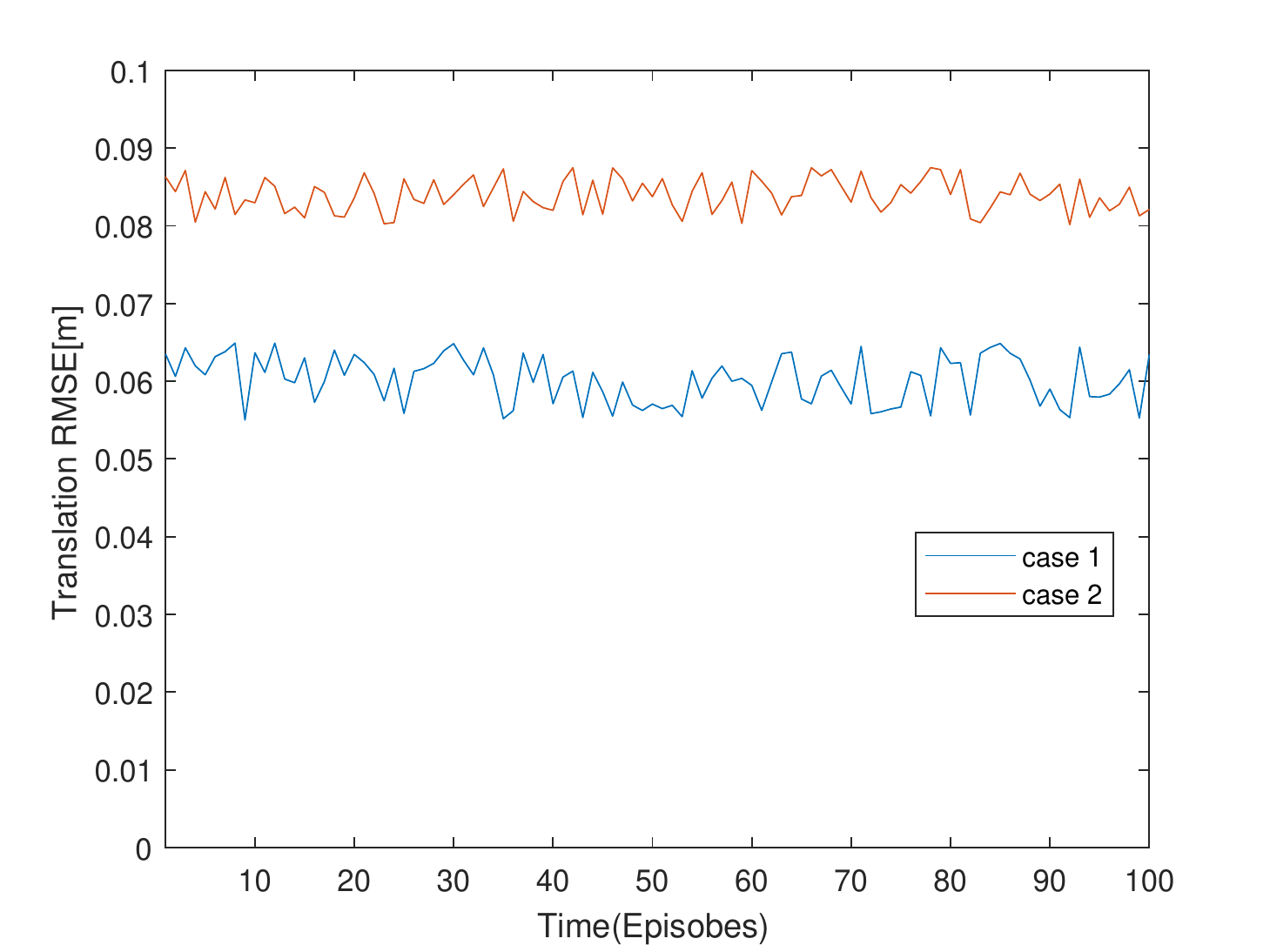}
  \includegraphics[width=0.45\linewidth]{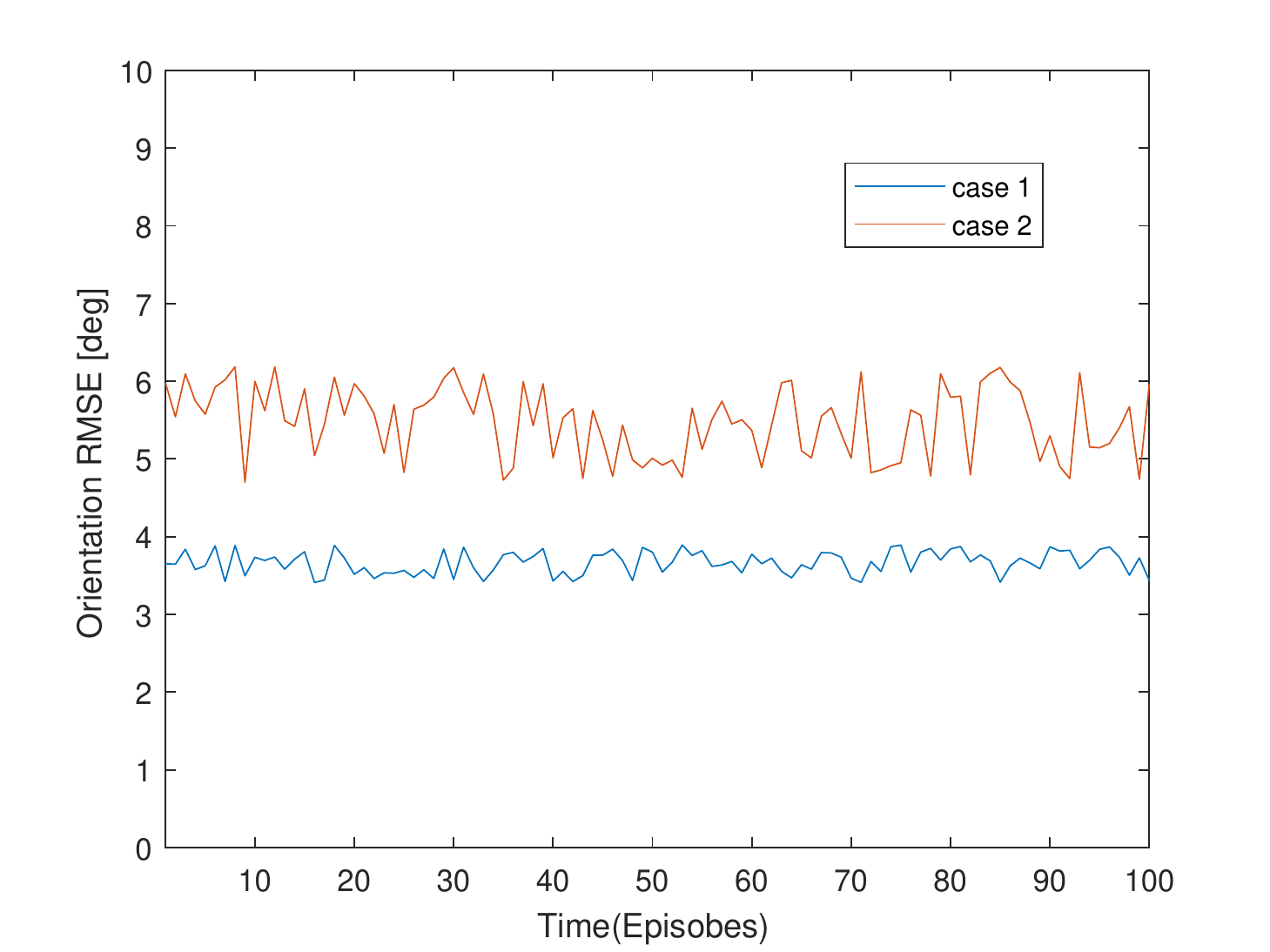}
  \caption{Result of experiment 2}
  \label{withoutfeature}
\end{figure}
\paragraph Experiment 3: This experiment is used to compare the proposed algorithm with other task allocation algorithms that take agent attributes as constants.
\begin{itemize}
  \item Case 1: The proposed collaboration mechanism based on MAS-DQN.
  \item Case 2: The proposed collaboration mechanism based on emotional recruitment model.
  \item Case 3: The proposed collaboration mechanism based on auction model.
\end{itemize}
\begin{figure}[H]\centering
  \includegraphics[width=0.45\linewidth]{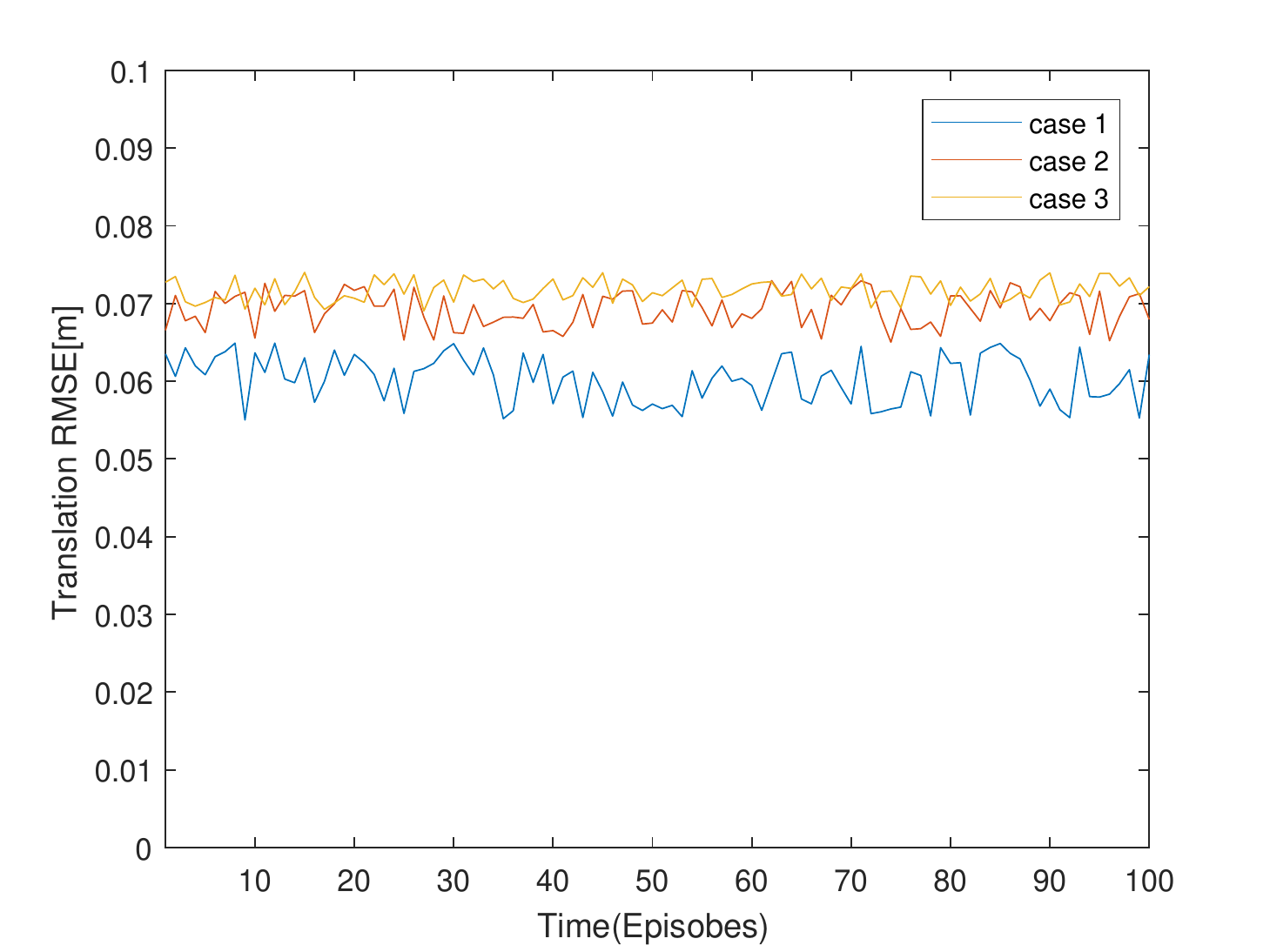}
  \includegraphics[width=0.45\linewidth]{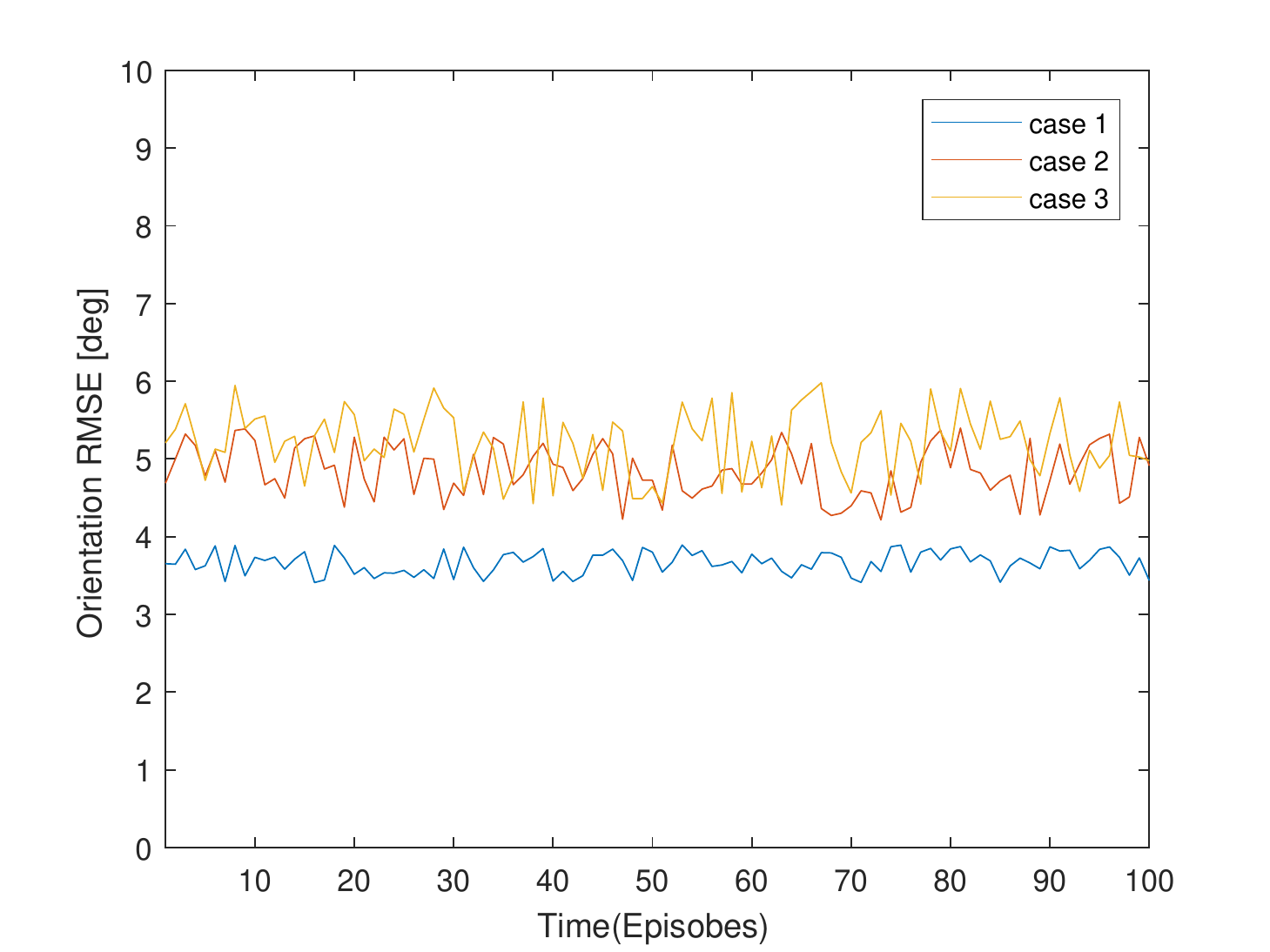}
  \caption{Result of experiment 3 }
  \label{withfeature}
\end{figure}
The proposed algorithm reduce transition RMSE(m) by 13.00\% and orientation position RMSE(deg) by 24.42\% compared with the mechanism based on emotional recruitment model. Compared with auction model, The MAS-DQN reduced transition RMSE by 16.03\% and orientation RMSE by 29.64\%.
\paragraph Experiment 4:In order to test the effectiveness of our algorithm in dealing with different degrees of difference in agent attributes, experiment 4 were performed when variance $\sigma_1$ was set to different values. The result of comparative experiment is shown in table \ref{tranwithsigma} and table \ref{ortawithsigma}. The proposed method reduces the average growth rate of transition RMSE by 16.58\% and 26.10\% by compared with auction model and emotional recruitment model.
The average growth rate of transition RMSE is reduced by 11.85\% and 8.83\%.

\begin{table}[htbp]
  \centering  
  \caption{robot transition RMSE(m) with different variance}  
  \label{tranwithsigma}  
  \begin{tabular}{|@{}l|c|c|r@{}|}
    \hline
    variance                    & 0.15   & 0.2    & 0.25   \\
    \hline
    Emotional recruitment model & 0.0638 & 0.0692 & 0.0752 \\
    \hline 
    Auction model               & 0.0657 & 0.0717 & 0.0761 \\
    \hline
    Proposed algorithm          & 0.0568 & 0.0602 & 0.0643 \\
    \hline
  \end{tabular}
\end{table}
\begin{table}[htbp]
  \centering  
  \caption{robot orientation RMSE(deg) with different variance} 
  \label{ortawithsigma}  
  \begin{tabular}{|@{}l|c|c|r@{}|}
    \hline
    variance                    & 0.15  & 0.2   & 0.25  \\
    \hline
    Emotional recruitment model & 4.689 & 4.849 & 5.620 \\
    \hline 
    Auction model               & 4.792 & 5.208 & 5.776 \\
    \hline
    Proposed algorithm          & 3.326 & 3.664 & 3.928 \\
    \hline
  \end{tabular}
\end{table}

\section{Conclusion}
This paper proposes a novel centralized multi-agent cooperative SLAM mechanism with considering the status and attributes of agents. The algorithm applies a new observation function based on ORBSLAM to apperceive the SLAM state of the whole MAS, a rainbow-based deep reinforcement learning framework called MAS-DQN is designed to learn the overall utility function $U(S,A(x))$ of the system, and the effectiveness of the framework is proved. The simulation results show MAS-DQN reduce transition RMSE(m) by 13.00\%, 13.06\% compared with emotional recruitment model and auction model, the orientation RMSE(deg) is also reduced by 24.42\%,29.64\% which is considered as great improvement on previously researched feature-based approaches. This algorithm inescapable brings huge computational cost to organizer, which makes the real-time of the system unable to be guaranteed. In the future, we will focus on distributed learning system to reduce the burden.

\end{document}